\documentclass[letterpaper, 10 pt, conference]{ieeeconf}  

\IEEEoverridecommandlockouts                              

\usepackage[pdftex]{graphicx}
\graphicspath{{Images/}}
\DeclareGraphicsExtensions{.pdf,.jpeg,.png}
\usepackage{amsmath} 
\usepackage[caption=false]{subfig}
\usepackage{multirow}
\usepackage{hyperref}

\title{\LARGE \bf City-Scale Road Audit System using Deep Learning}

\author{Sudhir Yarram \qquad Girish Varma \qquad C.V. Jawahar
\thanks{The authors are with Centre for Visual Information Technology,
Kohli Center for Intelligent System, IIIT Hyderabad.}\thanks{ Author's Email Addresses : yarramsudhirkumar.reddy@gmail.com,   girish.varma@iiit.ac.in,  jawahar@iiit.ac.in}%
\thanks{For project page containing code and dataset see: \url{https://cvit.iiit.ac.in/research/projects/cvit-projects/city-scale-road-audit}}
}

\begin{document}

\maketitle
\thispagestyle{empty}
\pagestyle{empty}

\begin{abstract}
Road networks in cities are massive and is a critical component of mobility. Fast response to defects, that can occur not only due to regular wear and tear but also because of extreme events like storms, is essential. Hence there is a need for an automated system that is quick, scalable and cost-effective for gathering information about defects. We propose a system for city-scale road audit, using some of the most recent developments in deep learning and semantic segmentation. For building and benchmarking the system, we curated a dataset which has annotations required for road defects. However, many of the labels required for road audit have high ambiguity which we overcome by proposing a label hierarchy. We also propose a multi-step deep learning model that segments the road, subdivide the road further into defects, tags the frame for each defect and finally localizes the defects on a map gathered using GPS. We analyze and evaluate the models on image tagging as well as segmentation at different levels of the label hierarchy.
\end{abstract}

\section{Introduction}
Cities across the globe are growing incredibly fast, both in area as well as population. Road network is a very important component of a city and any small disruption like traffic jams have a high cost in terms of safety, efficiency, and quality of life of citizens. Over time, even well-built roads degrade to form defects due to regular wear and tear, or even due to dynamic weather conditions like rain or storm. So, a process for frequent monitoring and identification of defects is required to resolve them. We call this process the ``road audit".  However, the sheer scale of the problem rules out any manual intervention at the reporting stage. An automated system that is cost-effective, scalable and easy to implement is essential. 

Processing images and inferring insights has been a well-known task in computer vision. The use of computer vision methods for road inspection was first proposed by \cite{varadharajan2014vision}. Later, deep convolutional neural networks were used to build a classifier for classifying image patches as 'crack' or 'non-crack' in \cite{zhang2016road}. Since then, the problem of semantic segmentation has seen significant improvements in performance due to deep learning based methods like Fully Convolutional Network ~\cite{long2015fully} and availability of large datasets for autonomous navigation \cite{Cordts_2016_CVPR,geiger2013vision}. These models and data sets enabled training versatile image feature representations for road scenes. However, autonomous navigation datasets only have a generic road label that needs to be further finely classified into the various road defects for the purpose of road audit. Many of the road defects like potholes, water logging can be ambiguous due to subtle differences in the texture of the surface and hence require more carefully designed methods.

   \begin{figure}[t]
   \vspace{0.7em}
\centering
      \includegraphics[width=1.0\linewidth] {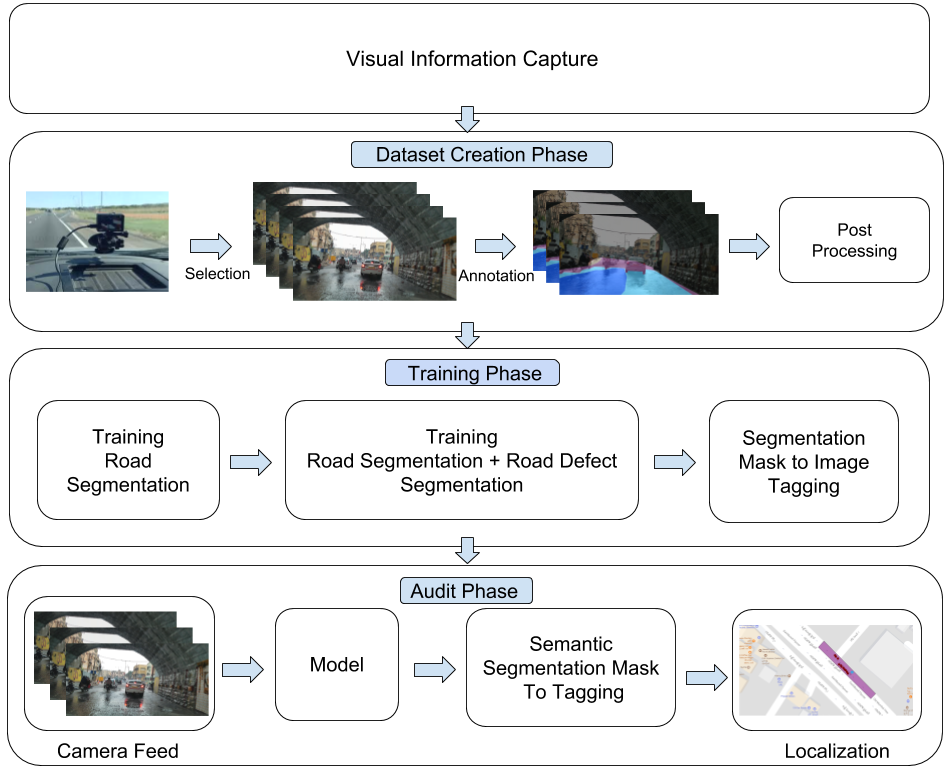}
      \caption{The four stages of our system: i) Visual Information Capture ii) Dataset Creation iii) Model Training iv) Audit (or Testing) Phase.}
      \label{fig:systemsetup}
       \vspace{-2.0em}
   \end{figure}

We build a cost-effective city-scale road audit system using some of the most recent developments in deep learning. Our system uses uncalibrated camera setup to capture visual information that can be mounted on vehicles that normally ply the roads compared to sophisticated setups required for autonomous navigation (see Section \ref{sec:camera}). We present a road-audit dataset collected from highly unstructured driving environments having annotations specific to the road audit requirements (see Section \ref{sec:dataset}). To minimize ambiguity among labels, we design a label hierarchy. We extract the relevant information about road quality from the dataset by modeling the problem as a cascaded semantic segmentation problem (see Section \ref{sec:approach}). This is followed by tagging each frame for defects using the semantic segmentation maps (see Section \ref{sec:audit-phase}). We analyze and evaluate the models on image tagging as well as segmentation at different levels of the label hierarchy (see Section \ref{sec:results}). Finally, we localize the defects and plot them on a map according to severity.  We ensure that all processing steps can be implemented in real time and be installed with little effort in a normal vehicle.  Hence using our system, a few vehicles can do a city-wide road audit greatly decreasing the man-hours required. 

\begin{figure*}[t]

\includegraphics[width=1.0\linewidth]{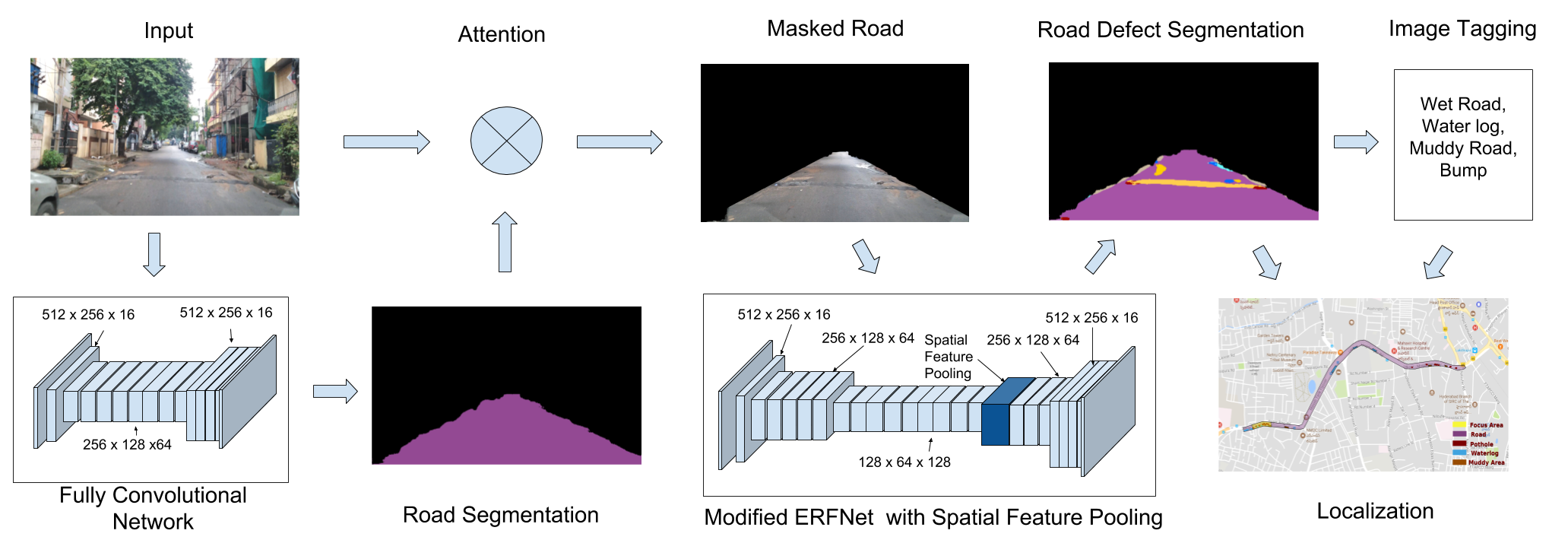}

\caption{A high-level overview of our proposed system. It is a multi-step procedure that uses more recent advances in real-time semantic segmentation. We segment the road, use it as an attention mechanism for obtaining road pixels, then use a powerful semantic segmentation model to segment multiple road defects. Finally, we tag the image frame for the defects and localize it in the map using GPS.}
\label{fig:csfpn}

\end{figure*}
\pagebreak
\section{Related Works}\label{sec:rel-works}
Road audit is mostly conducted by manual inspectors. This could also be crowdsourced to volunteers. Some of the initial automated systems used GPS and accelerometer only, followed by computer vision techniques for road audit.

\subsection{ Non-Computer Vision methods.}  
A system for pothole detection was proposed using just vibration and GPS sensor in ~\cite{potholepatrol} and using smartphone sensors like accelerometer and GPS in ~\cite{Wolv}. These sensors are highly error-prone and fail to detect other road defects like water-logging, muddy roads, and even minor rough patches that cannot be easily inferred from them. While our system depends on more reliable visual information with robust deep learning techniques and also caters to detecting more defects than just a pothole.

\subsection{Computer Vision methods.}  The first paper to propose the use of computer vision techniques for road inspection was \cite{varadharajan2014vision}. They used superpixel based segmentation techniques to identify road distress mostly restricted to cracks, in highly organized road conditions such as in a developed countries.  Image-based road distress detection was surveyed in \cite{wang2011automated}. Various crack segmentation methods are compared in \cite{kaul2010quantitative}. The community’s interest in crack detection has been growing recently \cite{DBLP:conf/itsc/SalmanMKR13,DBLP:journals/tits/MathavanKR15,DBLP:conf/icip/MedinaFCG14}. An integrated system for crack detection is proposed in \cite{oliveira2013automatic}. A CNN based neural networks to detect cracks was proposed by \cite{zhang2016road},     \cite{quintana2016simplified}. 


All these efforts were more focused on detection of cracks, potholes, and are more specialized for a specific region of roads, making it less adaptable to other, mostly unstructured regions. Most cities outside the developed world suffer from a much more diverse set of defects like water logging, rough roads, wet roads, muddy roads, etc. These defects are in some sense similar to cracks but they are more complex, ambiguous and the number of different defects makes it a challenging problem. Also, most of the previous methods use handcrafted features and traditional classifier like SVM with some using deep learning techniques for limited tasks. In this paper, we try to address more generic setting of road audit in unstructured road conditions. We also leverage the  advancements in real-time semantic segmentation and autonomous navigation datasets  to build an end-to-end deep learning model.

\section{Approach}\label{sec:approach}
We now describe our proposed system. Our system has four phases: i) Visual Information Capture ii) Dataset Creation iii) Model Training iv) Audit (or Testing) Phase as shown in Figure \ref{fig:systemsetup}. Dataset creation phase will be described in Section \ref{sec:dataset}. We describe the rest of the phases as well as the details of the deep learning model architecture in this section.

\subsection{Phase I: Visual Information Capture}\label{sec:camera}


Our system uses videos/images captured using a camera mounted on vehicles for information gathering. We achieve this by using a single uncalibrated camera setup, that could even be a GPS enabled smartphone camera, making it cost-effective and scalable. The camera is mounted inside the vehicle so that there are lesser chances of exposure to climatic conditions, ambient dust and also to avoid accidental loss of the camera. Using this setup, the system can be implemented with little or no additional cost making it cost-effective and scalable. 

\subsection{Phase II: Model Training}

We model the problem of localizing the road defects as a cascaded semantic segmentation problem, followed by image tagging and localization. All models and algorithms are chosen such that the system can be run in real-time at audit phase. This is required since the information gathering method described above generates a huge amount of data and it is not easy to store it and process it later. So our model focuses on extracting the required information in real time. It also captures the fine-grained information necessary to get statistics of road defects that are further localized.

\paragraph{Road Segmentation} As our primary goal is to segment road defects, the context extracted from dynamic and ambient objects like other vehicles, the sky, buildings, pedestrians etc. provides very little additional information for the model to segment road defects. Moreover, these might mislead the model. So, as the first stage, we segment road pixels from the background. As shown in Section \ref{sec:expres}, this proves to be effective. 


\paragraph{Fine-grained Segmentation of Roads}
Traditional semantic segmentation datasets like Cityscapes \cite{Cordts_2016_CVPR}, KITTI \cite{geiger2013vision}  have the notion of objects, which implies that they are reasonably well defined in the space of shape and color. However, as part of our road audit dataset, we have to deal with defects which are agnostic to the notion of shape, in a way, which can only be semantically differentiated through texture.
Incorporating these cues in the architecture as a second step, we further segment the road pixels into more fine-grained labels.


 
 



\subsection{Real-time Model Architecture}
In this section, we describe the different deep learning models used for solving this problem. We use a baseline network, as well as a custom designed network that incorporates the road segmentation and fine-grained defect segmentation into the network architecture. In the interest of creating a practical application, we have considered various models. Considering a right balance between efficiency and speed, we chose ERFNet~\cite{erfnet} (69.8\% with 41fps on Cityscapes) as our base-model. 

 With ERFNet as base-model, we experiment with two models: i.) fine-tuning a pre-trained cityscape model on our dataset ii.) train the model from scratch on our dataset. Note that both these models are initialized with the image features trained for Imagenet classification. ERFNet was designed for the application of autonomous navigation. Hence we also propose a modified architecture, which is more suited for the task of road defect segmentation.

Our proposed model is called the refined ERFNet, that contains a Cascade module and  Spatial Feature Pooling module on top of ERFNet. These modules incorporate the road segmentation as well as the fine-grained defect segmentation into the model architecture as shown in Figure \ref{fig:csfpn}. As an overview, our network takes an image as an input and outputs a semantic mask of the road and road related defects.
Our model works in multiple stages with Cascade Module and Spatial Feature Pooling Module designed specifically for the problem.



\subsubsection{Cascade Module} 
Cascade module consists of two submodules - road segmentation module and road defect segmentation modules sequentially. The task of the road segmentation module is to segment out road pixels from the background (non-road) pixels. For this, we use shortened ERFNet, trained on cityscapes and fine tune it to segment road pixels.  Then, the segmented mask generated is used to pass input road pixels to the second module using the attention mechanism. These road pixels are run through road-defect segmentation module, which further segments out road defects. These two models are trained in an end-to-end procedure.

\subsubsection{Spatial Feature Pooling Module}
The Spatial Feature Pooling Module aggregates the features for different scales of the image segments and tries to capture texture level cues. This caters to different scale variation in the road defects. 

\subsection{Phase IV : Audit} \label{sec:audit-phase}
In the audit phase, we deploy the real-time cascaded semantic segmentation model on a computer installed in the vehicle. It generates segmentation masks for the frames captured from the camera and does the following sequentially: 

\subsubsection{Image Tagging} After inferring the segmentation output from the input image, we decide on the appropriate labels for an image by thresholding the number of pixels segmented as that label. The threshold is decided by searching in the hyperparameter space on the validation data of the dataset.

\subsubsection{Localization}
The final stage of our system is to plot the road defects according to their severity on a map. We use the GPS information captured along with the image frame for this purpose. To indicate the severity, we use the number of pixels in the image classified as that particular label. However, doing this alone might not capture the severity, since a large defect could occupy only a small region due to viewing angles or even other vehicles blocking the view. Hence the severity score also needs to be averaged across nearby frames for getting a more accurate estimate.


\section{Dataset}\label{sec:dataset}


In this section, we describe the dataset creation phase of our system. Our motto behind the methodology of constructing the dataset was to capture very complex real-world scenes in an even more generic way. We collected data from many drive sequences (distinctive captures) spanning close to 60 kilometers, out of which we have selected around 18 driving sequences. These drive sequences cover a variety of highway roads, flyovers, and street roads. It covers a multitude of scenarios, right from having differently sized defects to various combination of defects in a single image. The dataset also incorporates a variety of traffic scenarios with distinct levels of diversity in terms of vehicles covering from sparsely crowded to densely crowded. It also incorporates diverse road settings, with partial to complete occlusions of the various road defects. In order to cover seasonal defects like water logging and wet roads, we had driven just after the rainy days.

In the frame selection process, we have ensured that we have taken well-curated images in terms of avoiding any kind of glare due to reflection. In order to counter the motion blur induced by residual vibration of the vehicle and due to sudden movements of the vehicle, we have carefully selected images that had blur below a threshold.

We sample frames from each of the sequences. Our sampling was driven by the idea to cover a wide variety of road defects. We annotate them coarsely.
The dataset has images of size 1024 $\times$ 2048 px. This dataset segments various semantic defects a road can have.

\subsubsection{Labels}\label{sec:labels} After careful collection of data, we have analyzed the kinds of road defects and have come up with tar road, pothole, water log, muddy road, obstruction, wet road, shoulder, cement road, bump labels. These labels are grouped in a three-level hierarchy. At the top, we have road and road-defects. Road is branched into class level labels namely cement road, tar road and shoulder labels and the rest go into road-defects. Then, road defects are subdivided into four categories category 1, category 2, category 3, category 4. Category 1 is subdivided into pothole, water log and wet road class labels, category 2 contains muddy road label. category 3 contains rough road label while category 4 is subdivided into obstructions and bump.

\begin{figure}[t]
\centering
\includegraphics[width=1.0\linewidth]{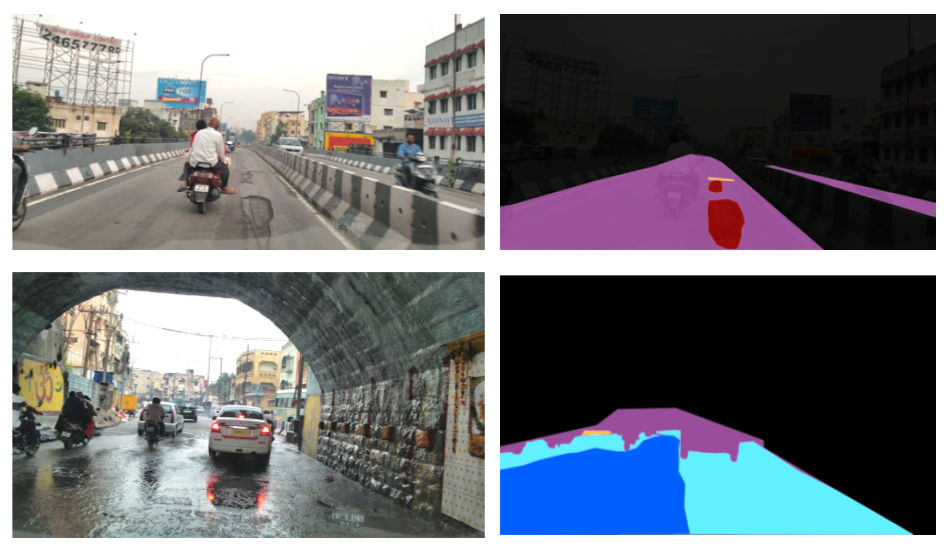}

\vspace{-1em}
   \caption{ The urban road is segmented for road defects. We can observe that road pixel annotation also includes pixels belonging to other objects like car, bike, etc., on the road which are later masked out as a post processing step.}
\label{fig:annot}
\label{fig:onecol}
\vspace{-1em}
\end{figure}

\subsubsection{Annotation Method}
We employ pixel-level annotation. We do this for two reasons 1) Annotation by image tagging is very subjective (How much area of a road image occupied by pothole be tagged as the pothole?). However, employing a supervised pixel level annotation allows annotators to mark all pixels that belong to label (say, pothole), thereby decreasing the subjectivity from one annotator to another. 2) Pixel level annotation provides more supervision for the model to learn better.
However, in the interest of resources, we have restrained ourselves to complete the task of annotation and basic quality control to about 15 minutes on average for a single frame. This is significantly less when compared to 45 minutes taken for annotation of typical urban road scene as reported by the cityscapes dataset \cite{Cordts_2016_CVPR} in their annotation protocol. Coming to our protocol, annotators were first asked to figure out the boundaries of road and shoulder. Then they annotate the road by ignoring all vehicles/objects/people present on the road. Later they segment the shoulder (or side-road). Then, they segment out road defects from already segmented road pixels circumventing the occlusions. The annotation was carried out in a coarse way; ensuring that all the essential pixels of a label are captured while loosening out the precision at label boundaries. Though this would hinder the performance of supervised algorithms, it was a necessary move, keeping in mind the sparsity of resources. Figure \ref{fig:annot} provides a sample annotation.
 
\subsubsection{Post Processing} In order to mask out other vehicle/object pixels included along with the road segments, we run a pre-trained cityscapes model on these images which segments other object pixels separated from road pixels. This generated cityscapes label mask and annotator generated annotation mask are combined to come up with a segmentation mask for the road and its defects excluding all other object pixels. Though this was not completely error-free, from our qualitative analysis, it was good enough for the task at hand. We have used cityscapes trained ERFNet for generating cityscape label mask and this is appropriate as the road pixel IoU was 97.9 which is close to state of the art (98.6) evaluated on cityscapes dataset.

\begin{figure}[t]
\vspace{-1em}
\centering
\includegraphics[width=1.0\linewidth]{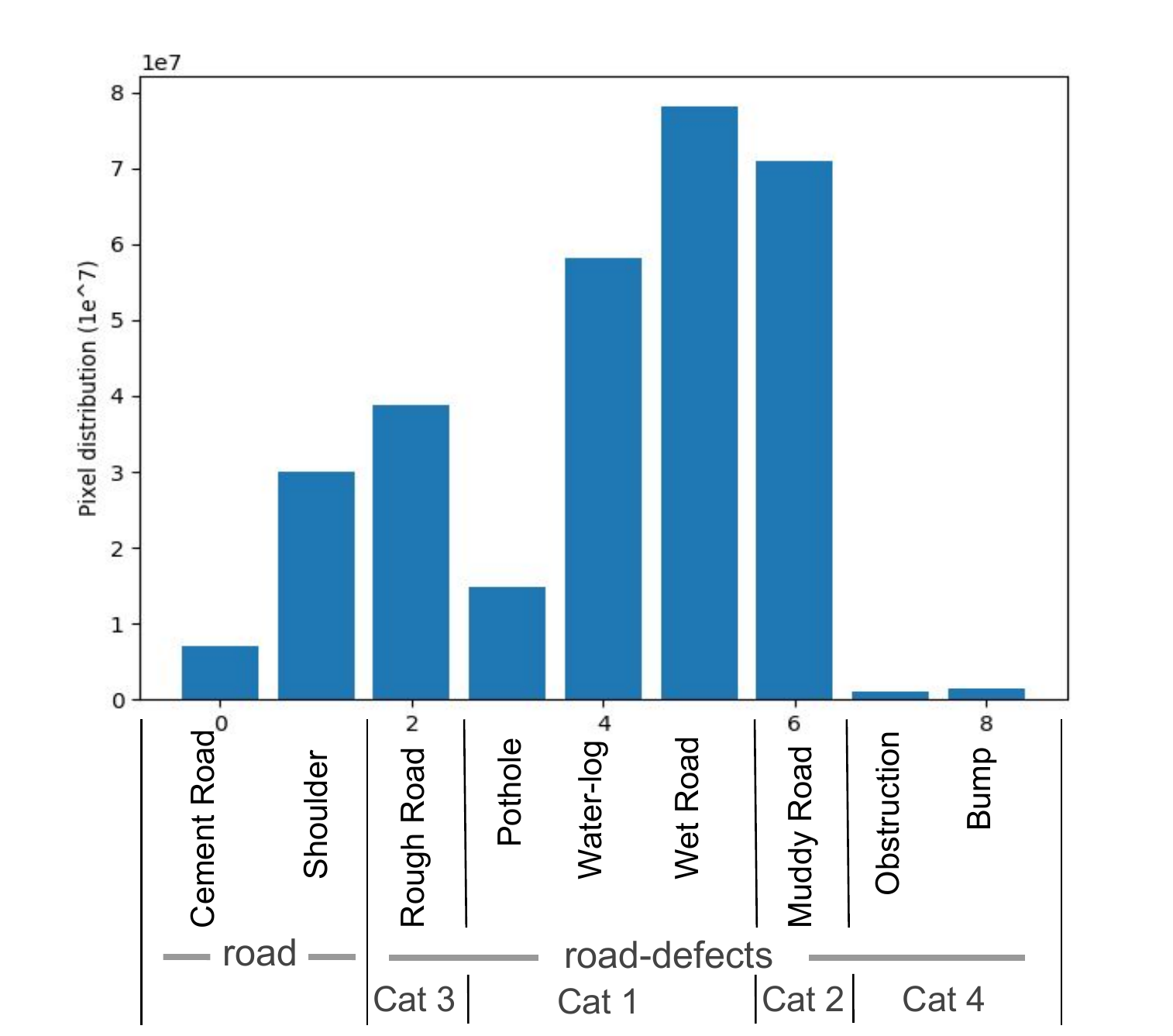}
\vspace{-1.5em}
   \caption{Here 'Cat' means Category. Pixel distribution for various defects.}
\label{fig:pixdis}
\label{fig:onecol}
\vspace{-1em}
\end{figure}

\begin{figure*}[t]
\centering
\includegraphics[width=1.0\linewidth]{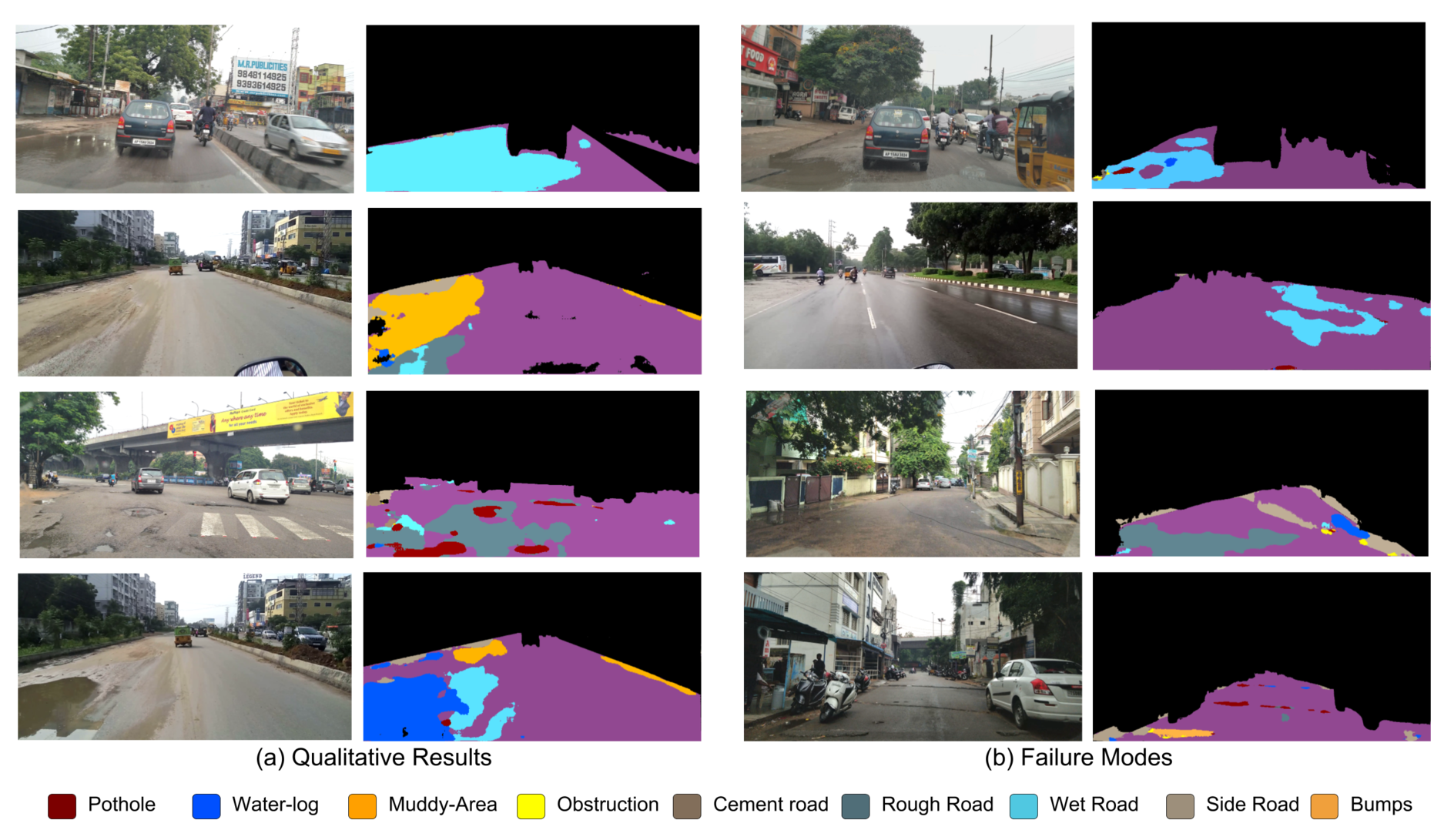}

   

\vspace{-0.8em}
\caption{(a) Qualitative Results for class level segmentation. These figures show the input image along with its segmentation mask as predicted by the refined ERFNet model. (b) Examples of Failures. Observations in the second column from top to bottom 1) Waterlog is segmented as wet road, 2) Only part of the wet road is segmented. 3) road is segmented as rough road 4) Bumps are not properly segmented.}
\label{fig:qualres}
\vspace{-0.8em}
\end{figure*}

\subsubsection{Label Statistics}

 The pixel distribution for each label in the train and in the test are ensured to be almost the same. Dataset labels pixel distribution is provided in Figure \ref{fig:pixdis}. The labels and the coarse annotation are sufficient in detecting the road defects in a satisfactory manner for the following two reasons. The labels that were incorporated cover all the major issues, which are generally reported in various surveys. The coarse annotation works for the task at hand as our primary focus is to detect the defects, which does not require pixel-level precision.

\section{Experiments}\label{sec:expres}


For all our experiments we have used the newly collected Indian Road dataset that was described in Section \ref{sec:dataset}. 
We train three semantic segmentation networks: ERFNet from scratch, fine-tuning cityscapes pre-trained ERFNet and refined ERFNet. We consider the first 2 as our baseline and compare them against the refined ERFNet.

\paragraph{Refined ERFNet} \label{sec:our-model}

We integrate the proposed Cascaded Spatial Feature Pooling module on the ERFNet architecture. We first train the initial module that segments out road pixels. We generate ground truth mask for road pixels as the union of ground truth of all the road defects. We train this module until the accuracies reach a satisfactory level. On top of this, we now attach a module whose task is to segment out road defect pixels. This module has an additional spatial feature pooling module. The generated output is compared with road defect ground truth to generate loss that is back-propagated.

We used the publicly available ERFnet code. 
It replaces the bottleneck module in the Resnet \cite{he2016deep} with more efficient and better-performing 1D-non-bottleneck module. We modified this network by adding the cascade module along with the spatial feature pooling module. For training, we have an input image of size 1024 $\times$ 512. We have used Adam optimizer with a learning rate of 5e-4 for a batch size of 14.



 
\subsubsection{Metrics} Since we are modeling the problem as a semantic segmentation problem, we use Intersection over Union (IoU) scores as accuracy metric. We evaluate the mean Intersection-over-Union. 
We also report the weighted mean Intersection over Union (wIoU) score. As pixel distribution among labels varies a lot, wIoU would give a good measure on how the algorithm is performing for all the pixels overall.
 
\subsubsection{Training complete setup end to end}
As described above, our model is trained end to end with a two-step training procedure. To gain more insights into the training procedure, we have independently trained the road level segmentation and defect level segmentation. This is contrasted with training in a two-step manner by first training road segmentation module to a satisfactory level and then attaching road-defect module and further train both together. We observed that the latter approach was better, resulting in a faster network convergence with similar performance.

\begin{table}
\caption{ Mean IoU and Inference speed on the test set of Indian Road dataset for category level and class level for three segmentation networks (Table Ia). (Table Ib) IoU scores computed for different levels in Dataset Hierarchy using our refined ErfNet. Here 'cat4' means category 4 (Table Ic) Iou scores for Individual classes using refined ErfNet. }

\centering

\subfloat[]{\begin{tabular}{|l|c|c|c|}
\hline
Method (Category Level) & IoU (\%) & wIoU (\%) & fps (1024 $\times$ 512)\\
\hline\hline
ERFNet (Fine-tuned) & 41.8 & 59.8 & 41.7\\
ERFNet (Scratch) & 42.1 & 59.8 & 41.7\\
Our refined ERFnet & 45.7 & 60.7 & 26.2 \\
\hline
\hline
Method (Class Level) & IoU (\%) & wIoU (\%) & fps (1024 $\times$ 512)\\
\hline
ERFNet (Fine-tuned) & 24.1 & 55.2 & 41.7\\
ERFNet (Scratch) & 24.4 & 57.2 & 41.7\\
Our refined ERFnet & 28.1 & 59.0 & 26.2 \\
\hline
\end{tabular}}
\vspace{0.5em}
\subfloat[]{\begin{tabular}{|l|c|}
\hline
Hierarchy(Our Method)  & IoU(\%) \\
\hline\hline
Road-defects & 65.1 \\
Category level & 45.7 \\
Class Level(-cat4) & 38.0\\
Class Level & 28.1 \\
\hline
\end{tabular}}
\hspace{0.1cm}
\subfloat[]{\begin{tabular}{|l|c|}
\hline
Our Method(Class) & IoU (\%) \\
   
\hline\hline
Tar Road & 72.5 \\
Pot hole & 23.5  \\
Rough Road & 34.5 \\
Cement Road & 13.8 \\
Obstruction & 10.7 \\
Wet Road & 32.3 \\
Water log & 28.8 \\
Muddy Road & 33.4 \\
Bumps & 03.3 \\
Shoulder & 28.1 \\
\hline

\end{tabular}}


\label{tab:miou}
\end{table}

\section{Results}\label{sec:results} The segmentation results of refined ERFNet and baseline are listed in Table \ref{tab:miou}. Refined ERFNet outperforms the original baselines by 3.6\%, 3.5\% for class level labels and category level labels respectively.The qualitative results are provided in Figure \ref{fig:qualres}. As present in Table \ref{tab:miou}, refined ERFNet gives a mean Intersection over union score of 65\% for (road vs defects), 45.77\% for category level (4 labels), 38.00\% for class level (8 labels  excluding category 4) and 28.13\% for class level (10 labels). The significant drop in accuracy from category level to class can be attributed to fine-grained separations among labels making it hard for the model. To some extent, the visual complexity in differentiating these labels was unavoidable during annotation. However, the municipal authorities can decide on what level of label hierarchy would suit their task and work accordingly. Our model infers at a speed of 26.2 fps on the Titan X GPU for an image of size 1024 $\times$ 512. Per class IoU scores are also provided in Table \ref{tab:miou}.
 
 


\paragraph{Cascade Module and Spatial feature pooling Module} We have experimented with a hierarchical segmentation of segmenting road pixels followed by road defects. Experiments show a boost in IoU scores by 3.6\%, 3.5\% for class and category levels respectively. This validates that our model of first segmenting out road pixels (which predominantly uses object cues like shape, color) and then segmenting out road defects is well suited for this task. It can also be inferred that our spatial feature pooling module is better at capturing the essential information necessary for segmenting out the fine-grained pixel differences among labels. 


\begin{table}
\caption{ Our experiment of tagging each image and thresholding the generated segmentation mask has given an F1 score of 50.3\% }
\centering
\begin{tabular}{|l|c|c|}
\hline
Our Method (Image Tagging) & Precision (\%) & Recall (\%) \\
\hline\hline
Pot hole & 59.3 & 83.3  \\
Rough Road & 67.7 & 61.7\\
Cement Road & 6.2 & 25.0 \\
Obstruction & 6.6 & 100.0 \\
Wet Road &  81.3 & 53.3\\
Water log & 56.2 & 54.5 \\
Muddy Road & 74.1 & 76.6 \\
Bumps & 28.5 & 50.0 \\
Shoulder & 16.0 & 25.0 \\
\hline

\end{tabular}
\label{tab:tagging}
\vspace{-2.5em}

\end{table}

\paragraph{Image Tagging} As our primary objective is to tag an image with road defects. We conducted an experiment on how would the model perform for an image tagging task. As ground truth, we asked the annotators to tag each image with all the class labels. Now, after generating a segmentation mask, we threshold the number of pixels belonging to each class and accordingly tag image with the corresponding label. With this, we achieve an F1 score of 50.3\% as shown in Table \ref{tab:tagging}. This shows that the model is good enough to recognize appropriate labels for an image. This provides additional validation for the appropriateness of the model to be used for practical purpose.

\paragraph{Failure Modes} Figure \ref{fig:qualres} gives examples of the failure modes of our model. The primary reason for failure is the essential similarity among labels. For example, in the top right example, we see that a water logging but is mislabeled as a wet road. However, precisely identifying such fine-grained differences are beyond human skills as many factors like illumination, lighting, surface reflectance play a crucial role.  Other less common modes of failure are due to the color variations in the texture of road due to sediments, inconsistency in road material.



\section{Conclusion}
We propose an automated system for road audit in large and unstructured cities. Our system is designed to be cost-effective, scalable and real time. Our data collection can be done using smartphones mounted inside normal vehicles. We curate a dataset with the required annotations for identifying fine-grained road defects, in unstructured road conditions. We propose a cascaded semantic segmentation model that can be trained using the dataset for segmenting the road defects. Furthermore, we tag the image with the defects using the segmentation masks and finally localize them on a map. We analyze and evaluate our models and show that they give reasonable performance for the task at hand. We expect that works along similar lines can help urban authorities identify areas vulnerable to damage by analyzing the changes roads underwent over a period of time. It could also be extended to infer reasons for traffic congestion, accident-prone areas etc.
\addtolength{\textheight}{-12cm}   



\vspace{-1.0em}
{\small
\bibliographystyle{IEEEtran}
\bibliography{IEEEexample}
}

\end{document}